%% file: main.tex
\documentclass[10pt,twocolumn,letterpaper]{article}

\usepackage[pagenumbers]{cvpr}      

\input{preamble}

\definecolor{cvprblue}{rgb}{0.21,0.49,0.74}
\usepackage[pagebackref,breaklinks,colorlinks,allcolors=cvprblue]{hyperref}

\def\paperID{10950} 
\def\confName{CVPR}
\def\confYear{2026}

\title{TRIDENT: A Trimodal Cascade Generative Framework for Drug and RNA-Conditioned Cellular Morphology Synthesis}

\author{
    Rui Peng$^{1,2}\textsuperscript{\#}$~
    Ziru Liu$^5\textsuperscript{\#}$~
    Lingyuan Ye$^{6,7}$~
    Yuxing Lu$^{1}$~
    Boxin Shi$^{3,4}\textsuperscript{*}$~
    Jinzhuo Wang$^{1}$\textsuperscript{*}\\
    \small \textsuperscript{1} Department of Big Data and Biomedical AI, College of Future Technology, Peking University\\
    \small \textsuperscript{2} Center for BioMed-X Research, Academy for Advanced Interdisciplinary Studies, Peking University\\
    \small \textsuperscript{3} State Key Laboratory of Multimedia Information Processing, School of Computer Science, Peking University\\ 
    \small \textsuperscript{4} National Engineering Research Center of Visual Technology, School of Computer Science, Peking University\\
    \small \textsuperscript{5} Yuanpei College, Peking University\qquad \textsuperscript{6} School of Life Sciences, Tsinghua University\\
    \small \textsuperscript{7} Peking University-Tsinghua University-National Institute of Biological Sciences Joint Graduate Program (PTN), Tsinghua University\\
    \small{\texttt{\{pengrui, lzr, luyx\}@stu.pku.edu.cn~~~yely23@mails.tsinghua.edu.cn}}\\
    \small{\texttt{shiboxin@pku.edu.cn~~~wangjinzhuo@pku.edu.cn}}
}

\begin{document}
\maketitle
\begingroup
  \renewcommand\thefootnote{}
  \footnotetext{\# Equal contribution.\quad * Corresponding authors.}
  \addtocounter{footnote}{-1}
\endgroup

\input{sec/0_abstract}  

\input{sec/1_intro}

\input{sec/2_related_work}

\input{sec/3_Methods}

\input{sec/4_results}

\input{sec/5_conclusion_limitation}

{
    \small
    \bibliographystyle{ieeenat_fullname}
    \bibliography{main}
}

\input{sec/X_suppl.tex}

\end{document}

%% file: preamble.tex
\usepackage[utf8]{inputenc} 
\usepackage[T1]{fontenc}    
\usepackage{amsmath,amssymb,mathtools,amsthm} 
\usepackage{microtype}      
\usepackage{xcolor}         
\usepackage{graphicx}       
\usepackage{algorithm}      
\usepackage{algorithmicx}   
\usepackage{algpseudocode}  
\usepackage{multirow}       
\usepackage{adjustbox}      
\usepackage[table]{xcolor}  
\usepackage{booktabs}       
\usepackage{subcaption}

\newcommand{\E}{\mathbb{E}}
\newcommand{\KL}{\text{D}_{\text{KL}}}
\newcommand{\vect}[1]{\boldsymbol{#1}} 
\newcommand{\mat}[1]{\mathbf{#1}}     



%% file: sec/0_abstract.tex
\begin{abstract}
Accurately modeling the relationship between perturbations, transcriptional responses, and phenotypic changes is essential for building an AI Virtual Cell (AIVC). However, existing methods typically constrained to modeling direct associations, such as \textit{Perturbation $\rightarrow$ RNA} or \textit{Perturbation $\rightarrow$ Morphology}, overlook the crucial causal link from RNA to morphology. To bridge this gap, we propose TRIDENT, a cascade generative framework that synthesizes realistic cellular morphology by conditioning on both the perturbation and the corresponding gene expression profile. To train and evaluate this task, we construct MorphoGene, a new dataset pairing L1000 gene expression with Cell Painting images for 98 compounds. TRIDENT significantly outperforms state-of-the-art approaches, achieving up to 7-fold improvement with strong generalization to unseen compounds. In a case study on docetaxel, we validate that RNA-guided synthesis accurately produces the corresponding phenotype. An ablation study further confirms that this RNA conditioning is essential for the model's high fidelity. By explicitly modeling transcriptome–phenome mapping, TRIDENT provides a powerful in silico tool and moves us closer to a predictive virtual cell.
\end{abstract}

%% file: sec/1_intro.tex
\begin{figure}[t]
  \centering
  \includegraphics[width=\linewidth]{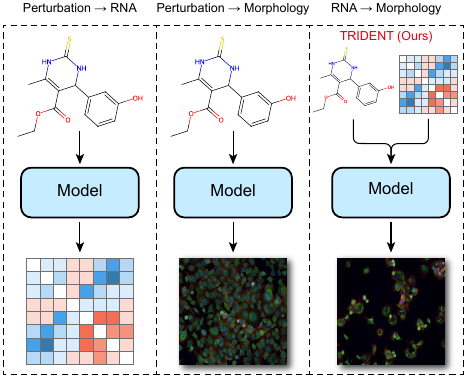}
  \caption{A comparison of cellular response modeling tasks. (Left) Predicting RNA from perturbation. (Middle) Predicting morphology from perturbation. (Right) Our model, TRIDENT, which integrates both perturbation and RNA to predict morphology, explicitly learning the \textit{RNA $\rightarrow$ Morphology} relationship.
  }
  \label{fig1}
  \vspace{-1.5em}
\end{figure}

\section{Introduction}
\label{intro}

High-throughput omics and artificial intelligence are advancing the vision of AI Virtual Cell (AIVC), a digital twin aiming to simulate cellular behavior across diverse states and scales \cite{bunne2024build}. Foundational to realizing this vision requires accurately modeling a cell's response to perturbations, which involves three key elements: the perturbation itself, the resulting gene expression (RNA) changes, and the ultimate phenotypic outcome in cellular morphology \cite{way2022morphology}.

While technologies like L1000 profiling \cite{subramanian2017next} and Cell Painting \cite{bray2016cell} have enabled models that predict \textit{Perturbation $\rightarrow$ RNA} \cite{qi2024predicting,adduri2025predicting,roohani2024predicting,hetzel2022predicting} or \textit{Perturbation $\rightarrow$ Morphology} \cite{palma2025predicting,navidi2025morphodiff, yang2021mol2image}, they overlook the fundamental \textit{RNA $\rightarrow$ Morphology} (\cref{fig1}). This gap limits our ability to simulate the virtual cell as an integrated system where molecular events mechanistically drive phenotypic outcomes. Meanwhile, generative models, particularly diffusion models \cite{ho2020denoising,nichol2021improved,dhariwal2021diffusion,ho2021classifier,nichol2022glide,songdenoising,peebles2023scalable} and VAEs \cite{kingma2013auto,higgins2017beta,van2017neural,mescheder2017adversarial,bao2017cvae}, have shown great potential in biological image synthesis and RNA reconstruction \cite{qi2024predicting,lotfollahi2019scgen,lotfollahi2020conditional,rampavsek2019dr,wu2025couplevae,doncevic2023biologically}. These advances demonstrate the potential of deep generative models to capture intricate biological patterns. However, within the specific context of virtual cell modeling, their application has not yet addressed the critical challenge of explicitly learning the cross-modal mapping from transcriptome to phenome.

To bridge this critical gap, we introduce TRIDENT (\textbf{TR}anscription-drug \textbf{I}nformed latent \textbf{D}iffusion \textbf{E}mbedding \textbf{N}e\textbf{T}work for cellular morphology synthesis), a cascade generative framework that models the complete perturbation-RNA-morphology relationship (\cref{fig1}). TRIDENT first uses a VAE to integrate drug and RNA profiles into a unified latent embedding, which then conditions a Diffusion Transformer (DiT) \cite{peebles2023scalable} to synthesize high-fidelity cellular morphology. This explicit \textit{(Perturbation + RNA) $\rightarrow$ Morphology} pathway enables a more mechanistic approach to virtual cell modeling. Our contributions are summarized as follows:
\begin{itemize}
    \item We introduce TRIDENT, a novel cascade generative framework that, to our knowledge, is the first to model the complete tripartite relationship between perturbation, RNA, and cellular morphology by explicitly learning the fundamental RNA $\rightarrow$ Morphology mapping.
    \item TRIDENT generates cellular morphologies with state-of-the-art fidelity, significantly outperforming existing models in both in-distribution and challenging out-of-distribution settings.
    \item We construct MorphoGene, a new, paired trimodal dataset that integrates gene expression and cellular morphology data for 98 small-molecule drugs.
\end{itemize}

%% file: sec/2_related_work.tex
\section{Related Work}
\label{related_work}

\noindent\textbf{Perturbation $\rightarrow$ RNA prediction.} Significant effort addresses the prediction of transcriptomic responses to genetic or chemical perturbations. These approaches employ architectures like graph neural networks (GEARS \cite{roohani2024predicting}), encoder-decoders (chemCPA \cite{hetzel2022predicting}, PRnet \cite{qi2024predicting}), and transformers (STATE \cite{adduri2025predicting}). A complementary line of VAE-based models (e.g., scGen \cite{lotfollahi2019scgen}, Dr.VAE \cite{rampavsek2019dr}, CoupleVAE \cite{wu2025couplevae}, OntoVAE \cite{doncevic2023biologically}) directly learns counterfactual gene expression. While effective at the \textit{Perturbation $\rightarrow$ RNA} mapping, these models stop short of linking molecular changes to downstream phenotypes.
\vspace{0.5em}

\noindent\textbf{Perturbation $\rightarrow$ Morphology Prediction.}
Predicting morphological responses is a growing application of conditional image generation, often leveraging diffusion models. Pioneering works generated histopathology from RNA (RNA-CDM \cite{carrillo2025generation}) or synthesized cellular structures \cite{moghadam2023morphology}. Current methods aim to simulate 2D or 3D post-perturbation phenotypes directly, including MorphoDiff \cite{navidi2025morphodiff}, IMPA \cite{palma2025predicting}, Mol2Image \cite{yang2021mol2image}, and DISPR \cite{waibel2023diffusion}. These approaches, however, typically bypass the intermediate molecular state. They do not explicitly model how transcriptional changes orchestrate morphological alterations, leaving the fundamental \textit{RNA $\rightarrow$ Morphology} mapping as a black box.

%% file: sec/3_Methods.tex
\section{Method}

\begin{figure*}[t]
  \centering
  \includegraphics[width=\linewidth]{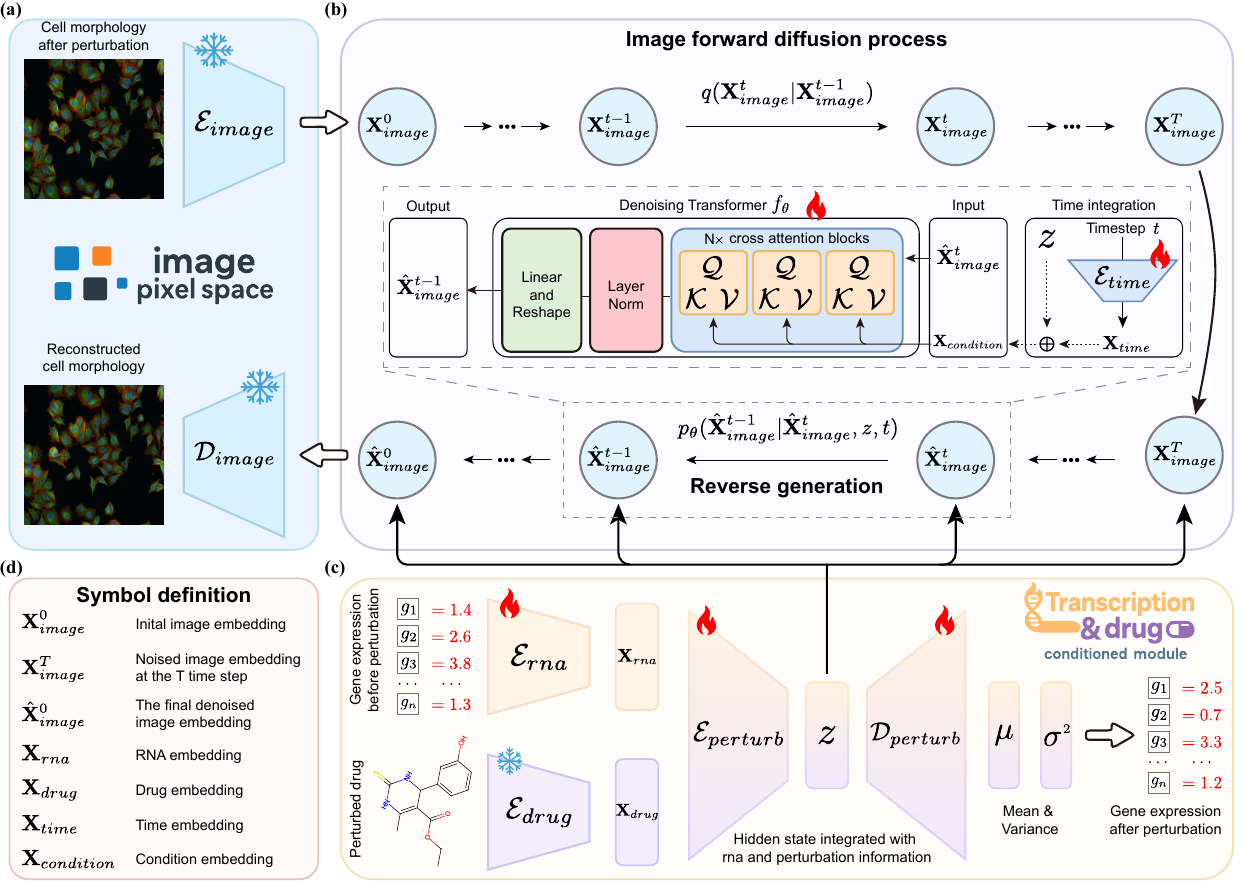}
  \vspace{-1.5em}
  \caption{Overview of the TRIDENT framework. (a) A VAE maps the high-resolution morphology images from pixel space into a compressed latent representation. (b) The Morphology Generation Module. A denoising transformer learns to reverse a forward noising process, using cross-attention to integrate a guiding condition vector that combines RNA-drug latent and time information. (c) The Transcription-Drug Condition Module. A VAE-based module encodes pre-perturbation gene expression and drug information into a latent vector, which is used to guide image generation. (d) Symbol definitions.
  }
  \label{fig2}
  \vspace{-1.5em}
\end{figure*}

TRIDENT models the mapping from gene expression to cellular morphology, conditioned on a drug perturbation $D$. We formally define this as learning the conditional probability $p\,(I \mid G_{pre}, D\,)$, where $I$ is the morphology and $G_{pre}$ is the pre-perturbation gene expression. As shown in \cref{fig2}, we use a two-stage cascade architecture to model this complex, cross-modal relationship. To train this model, we construct MorphoGene, a novel trimodal dataset integrating Cell Painting data from BBBC021 \cite{caie2010high} with L1000 gene expression profiles \cite{subramanian2017next}. We will sequentially describe the dataset construction, the training and inference process in the following sections.

\subsection{MorphoGene Dataset Construction}
MorphoGene integrates morphology from BBBC021 (MCF7 breast cancer cell line) and gene expression from L1000, linked by 98 small-molecule perturbagens. For morphology data, we merge the DAPI (blue), tubulin (green), and actin (red) channels into RGB composites and cropping them to 512x512. For gene expression, we averaged all corresponding L1000 profiles for each of the 98 compounds into a single representative vector. We then augmented the image collection for each compound to 1,000 samples, creating a total corpus of 98,000 trimodal samples.

We partitioned these 98 compounds to create training, in-distribution (ID), and out-of-distribution (OOD) test sets to evaluate the model's generalization capabilities:
\begin{itemize}
    \item \textbf{Training and ID Cohort:} 44 compounds present in both datasets. For each of these compounds, their samples were split 8:2 to form the training set and the ID test set.
    \item \textbf{OOD Cohort:} The remaining 54 compounds were held out entirely, with all associated samples forming the OOD test set.
\end{itemize}
Each sample in the final dataset contains the drug $D$, image $I$, and the corresponding averaged pre- and post-perturbation gene expression profiles $G_{pre}$, $G_{post}$.

\subsection{Transcription-Drug Condition Module}

The core function of this module is to compute a comprehensive conditional embedding $z$ that encodes both the cell's molecular state and the applied perturbation. As depicted in \cref{fig2}c, this module is structured as a VAE architecture where the encoder maps the inputs ($G_{pre}, D$) to a latent space, and the decoder reconstructs the outcome ($G_{post}$). Formally, the inputs $G_{pre} \in \mathbb{R}^{N_{genes}}$ and the drug's molecular representation $D$ (e.g., a SMILES string) are projected into their respective embedding spaces via dedicated encoders: $\mathbf{X}_{rna} = \mathcal{E}_{rna}(G_{pre}) \;,\; \mathbf{X}_{drug} = \mathcal{E}_{drug}(D)$.

These two embeddings are concatenated and processed by a perturbation encoder $\mathcal{E}_{perturb}$ to parameterize a posterior distribution $q_{\phi}(z | G_{pre}, D)$, which is modeled as a diagonal Gaussian:
\begin{equation}
[\mu_z, \log \sigma^2_z] = \mathcal{E}_{perturb}([\mathbf{X}_{rna}, \mathbf{X}_{drug}])\;,
\end{equation}
The latent vector $z$ is then sampled using the reparameterization trick: $z = \mu_z + \sigma_z \odot \epsilon_z$, where $\epsilon_z \sim \mathcal{N}(\mathbf{0}, \mathbf{I})$.

To ensure $z$ captures biologically salient information regarding the perturbation's effect, we regularize the latent space by forcing it to predict the resulting post-perturbation gene expression $G_{post}$. A decoder $\mathcal{D}_{perturb}$ models the likelihood $p_{\psi}(G_{post} | z)$, also as a Gaussian, outputting its parameters $\mu$ and $\sigma^2$:
\begin{equation}
[\mu_{G_{post}}, \log \sigma^2_{G_{post}}] = \mathcal{D}_{perturb}(z)\;,
\end{equation}

Finally, this module is trained by maximizing the Evidence Lower Bound (ELBO), which consists of a reconstruction term and a Kullback-Leibler (KL) divergence term. The corresponding loss function $\mathcal{L}_{VAE}$ is:
\begin{equation}
\begin{aligned}
\mathcal{L}_{VAE} = \underbrace{\mathbb{E}_{q_{\phi}(z | G_{pre}, D)}[-\log p_{\psi}(G_{post} | z)]}_{\text{Reconstruction Loss}} + \\
\underbrace{D_{KL}(q_{\phi}(z | G_{pre}, D) \,||\, p(z))}_{\text{Regularization (KL Divergence)}}\;,
\end{aligned}
\end{equation}
The prior $p(z)$ is a standard normal distribution $\mathcal{N}(\mathbf{0}, \mathbf{I})$. 

Minimizing this loss forces $z$ to be a compact representation that is predictive of the post-perturbation molecular state and retain the critical information linking the initial state ($G_{pre}$), the perturbation ($D$), and the resultant molecular outcome ($G_{post}$). This latent vector $z$, serving as the condition, is then passed to the Morphology Generation Module to guide the final image synthesis.

\subsection{Morphology Generation Module}

This module generates the high-resolution cell morphology $\hat{\mathbf{X}}^0_{image}$ conditioned on $\mathbf{X}_{condition}$. We employ the latent diffusion model (LDM) framework, performing the diffusion process within a compressed latent space for computational tractability.
\vspace{0.5em}

\noindent\textbf{Image Latent Space.} As shown in \cref{fig2}a, a pre-trained VAE, consisting of an encoder $\mathcal{E}_{image}$ and a decoder $\mathcal{D}_{image}$, is utilized. A high-resolution image $I$ is first encoded into a latent representation $\mathbf{X}^0_{image} = \mathcal{E}_{image}(I)$. The diffusion process operates entirely on $\mathbf{X}^0_{image}$. The final generated latent $\hat{\mathbf{X}}^0_{image}$ is then transformed back to pixel space via $\hat{I} = \mathcal{D}_{image}(\hat{\mathbf{X}}^0_{image})$.
\vspace{0.7em}

\noindent\textbf{Forward Diffusion Process ($q$).} Following \cref{fig2}b, the training process is anchored by a fixed forward diffusion process $q$, which incrementally corrupts the initial image latent $\mathbf{X}^0_{image}$ over $T$ discrete timesteps. This process is defined as a Markov chain that gradually adds Gaussian noise according to a pre-defined variance schedule $\{\beta_t \in (0, 1)\}_{t=1}^T$:
\begin{equation}
\resizebox{0.91\linewidth}{!}{$
q(\mathbf{X}^t_{image} \mid \mathbf{X}^{t-1}_{image}) =
\mathcal{N}(\mathbf{X}^t_{image}; \sqrt{1-\beta_t}\mathbf{X}^{t-1}_{image}, \beta_t \mathbf{I})\;,
$}
\end{equation}

A key property of this Markov chain is that we can sample the latent state $\mathbf{X}^t_{image}$ at any arbitrary timestep $t$ in a closed form, conditioned only on the initial state $\mathbf{X}^0_{image}$. Using the notation $\alpha_t = 1 - \beta_t$ and $\bar{\alpha}_t = \prod_{s=1}^t \alpha_s$, the distribution of $\mathbf{X}^t_{image}$ is given by:
\begin{equation}
\resizebox{0.91\linewidth}{!}{$
    q(\mathbf{X}^t_{image} \mid \mathbf{X}^0_{image}) = \mathcal{N}(\mathbf{X}^t_{image}; \sqrt{\bar{\alpha}_t}\mathbf{X}^0_{image}, (1-\bar{\alpha}_t)\mathbf{I})\;,
$}
\end{equation}

This allows us to directly generate a noisy sample for any timestep $t$ by sampling a standard Gaussian noise variable $\epsilon \sim \mathcal{N}(\mathbf{0}, \mathbf{I})$ and applying the reparameterization:
\begin{equation}
    \mathbf{X}^t_{image} = \sqrt{\bar{\alpha}_t}\mathbf{X}^0_{image} + \sqrt{1-\bar{\alpha}_t}\epsilon\;,
\end{equation}
This equation represents the explicit noising process, which provides the noisy inputs for training the model.
\vspace{0.7em}

\noindent\textbf{Reverse Denoising Process ($p_{\theta}$) and Objective Function.} The objective of training is to learn a reverse process $p_{\theta}(\hat{\mathbf{X}}^{t-1}_{image} \mid \hat{\mathbf{X}}^t_{image}, z, t)$ that can invert the diffusion, effectively learning to denoise the corrupted latents. This reverse process $p_{\theta}$ is parameterized as a Gaussian:
\begin{equation}
\mathcal{N}(\hat{\mathbf{X}}^{t-1}_{image}; \mu_{\theta}(\hat{\mathbf{X}}^t_{image}, z, t), \Sigma_{\theta}(\hat{\mathbf{X}}^t_{image}, z, t))\;,
\end{equation}
Following the DDPM framework \cite{ho2020denoising}, we set the covariance to untrained constants $\Sigma_{\theta} = \sigma_t^2 \mathbf{I}$, where $\sigma_t^2$ is typically set to $\beta_t$. The mean $\mu_{\theta}$ is parameterized to predict the noise $\epsilon$ that was added during the forward process. 

As shown in \cref{fig2}b, we implement this noise predictor as a denoising transformer $f_{\theta}$. The predicted mean $\mu_{\theta}$ is then derived from this noise prediction:
\begin{equation}
\resizebox{0.91\linewidth}{!}{$
  \mu_{\theta} = \frac{1}{\sqrt{\alpha_t}} \left( \hat{\mathbf{X}}^t_{image} - \frac{\beta_t}{\sqrt{1-\bar{\alpha}_t}} f_{\theta}(\hat{\mathbf{X}}^t_{image}, \mathbf{X}_{condition}) \right)\;,  
$}
\end{equation}
The network $f_{\theta}$ takes two inputs: (1) The noisy image latent $\hat{\mathbf{X}}^t_{image}$ and (2) The latent embedding condition embedding $\mathbf{X}_{condition}$.

As depicted in the Time integration block of \cref{fig2}b, the condition vector $\mathbf{X}_{condition}$ is formed by the element-wise addition of the latent vector $z$ (from the Transcription-Drug Condition module) and the time embedding $\mathbf{X}_{time}$, which is generated by passing the timestep $t$ through an embedding layer $\mathcal{E}_{time}$.

The denoising transformer $f_{\theta}$ consists of $N$ stacked blocks. Each block integrates the condition $\mathbf{X}_{condition}$ via a cross-attention mechanism, where the image representation provides the queries ($Q$) while $\mathbf{X}_{condition}$ provides the keys ($K$) and values ($V$). This repeated application of cross-attention throughout the network's depth is the critical mechanism that forces the model to learn the complex association between the RNA-drug condition and the morphological features. The model $f_{\theta}$ is then optimized via a simplified $L_2$ objective:
\begin{equation}
\mathcal{L}_{LDM} = \mathbb{E}_{t, \mathbf{X}^0_{image}, \epsilon, z} \left[ || \epsilon - f_{\theta}(\hat{\mathbf{X}}^t_{image}, \mathbf{X}_{condition}) ||^2 \right],
\end{equation}
where $\hat{\mathbf{X}}^t_{image} = \sqrt{\bar{\alpha}_t}\mathbf{X}^0_{image} + \sqrt{1-\bar{\alpha}_t}\epsilon$.

The final, end-to-end training objective for TRIDENT is the combined loss from both modules. We optimize the sum of the VAE loss and the LDM loss:
\begin{equation}
\mathcal{L}_{TRIDENT} = \mathcal{L}_{VAE} + \mathcal{L}_{LDM}\;,
\end{equation}

Optimizing this total loss jointly trains the framework to learn a conditional latent space $z$ that is both predictive of molecular outcomes ($G_{post}$) and informative for guiding the synthesis of high-fidelity morphological images ($I$). See the supplementary material for detailed pseudocode.

\subsection{Inference Procedure}
The inference process generates a novel cellular morphology $\hat{I}$ by reversing the diffusion process. The procedure begins by sampling an initial latent variable from the prior distribution $\hat{\mathbf{X}}^T_{image} \sim \mathcal{N}(\mathbf{0}, \mathbf{I})$. Given the pre-perturbation gene expression $G_{pre}$ and drug $D$, the Transcription-Drug Condition Module first computes the condition vector $z$.

The model then iteratively denoises the latent variable $\hat{\mathbf{X}}^T_{image}$ for $t = T, T-1, ..., 1$, using the learned conditional distribution $p_{\theta}(\hat{\mathbf{X}}^{t-1}_{image} \mid \hat{\mathbf{X}}^t_{image}, z, t)$. Each step of this reverse Markov chain computes a slightly less noisy latent $\hat{\mathbf{X}}^{t-1}_{image}$ from the previous $\hat{\mathbf{X}}^t_{image}$. Using our noise-prediction parameterization $f_{\theta}$, the update step is given by:
\begin{equation}
\resizebox{0.85\linewidth}{!}{$
\begin{aligned}
\hat{\mathbf{X}}^{t-1}_{image}
&= \frac{1}{\sqrt{\alpha_t}}\left( \hat{\mathbf{X}}^t_{image} - \right.\\
&\quad \left. \frac{\beta_t}{\sqrt{1-\bar{\alpha}_t}}
  f_{\theta}(\hat{\mathbf{X}}^t_{image}, \mathbf{X}_{condition})
\right) + \sigma_t \boldsymbol{\epsilon}'\;,
\end{aligned}
$}
\end{equation}
Here, the coefficients $\alpha_t$ and $\bar{\alpha}_t$ are derived from the fixed variance schedule $\{\beta_t\}_{t=1}^T$ of the forward process. $\mathbf{\epsilon}'$ is a random Gaussian noise sample $\mathbf{\epsilon}' \sim \mathcal{N}(\mathbf{0}, \mathbf{I})$ for $t > 1$, and $\mathbf{\epsilon}' = 0$ for $t=1$. 

After $T$ steps, this process yields the final denoised latent representation $\hat{\mathbf{X}}^0_{image}$. Finally, this latent representation is passed through the image decoder $\mathcal{D}_{image}$ to reconstruct the final high-resolution morphology image: $\hat{I} = \mathcal{D}_{image}(\hat{\mathbf{X}}^0_{image})$.

%% file: sec/4_results.tex
\section{Experiments}

\begingroup
\setlength{\tabcolsep}{4pt}
\renewcommand{\arraystretch}{1.35}
\begin{table}[t]
  \centering
  \caption{Quantitative comparison of TRIDENT against SOTA baselines on the in-distribution and out-of-distribution test sets. Performance is measured by FID, KID, and IS. Lower scores indicate better performance.}
  \label{table1}
  \begin{adjustbox}{width=\columnwidth}
  \begin{tabular}{l|ccc|ccc}
  \toprule\toprule
  \multirow{2}{*}{\centering\bfseries\large Methods} &
    \multicolumn{3}{c|}{\textbf{In-Distribution}} &
    \multicolumn{3}{c}{\textbf{Out-of-Distribution}} \\
  \cmidrule(lr){2-4}\cmidrule(lr){5-7}   
  & FID$\downarrow$ & KID$\downarrow$ & IS$\downarrow$ & FID$\downarrow$ & KID$\downarrow$ & IS$\downarrow$ \\
  \midrule\midrule
    MorphoDiff & 250.290 & 0.248 & 2.614 & 387.135 & 0.436 & 2.747 \\
    Stable Diffusion & 354.576 & 0.378 & 2.792 & 393.129 & 0.543 & 2.932 \\
    \cellcolor{gray!50}\textbf{TRIDENT (ours)} &
      \cellcolor{gray!50}\textbf{49.770} &
      \cellcolor{gray!50}\textbf{0.013} &
      \cellcolor{gray!50}\textbf{2.240} &
      \cellcolor{gray!50}\textbf{126.150} &
      \cellcolor{gray!50}\textbf{0.222} &
      \cellcolor{gray!50}\textbf{2.523} \\
    \bottomrule\bottomrule
  \end{tabular}
  \end{adjustbox}
  \vspace{-1.5em}
\end{table}
\endgroup 

\begin{figure*}[t]
  \centering
  \includegraphics[width=\linewidth]{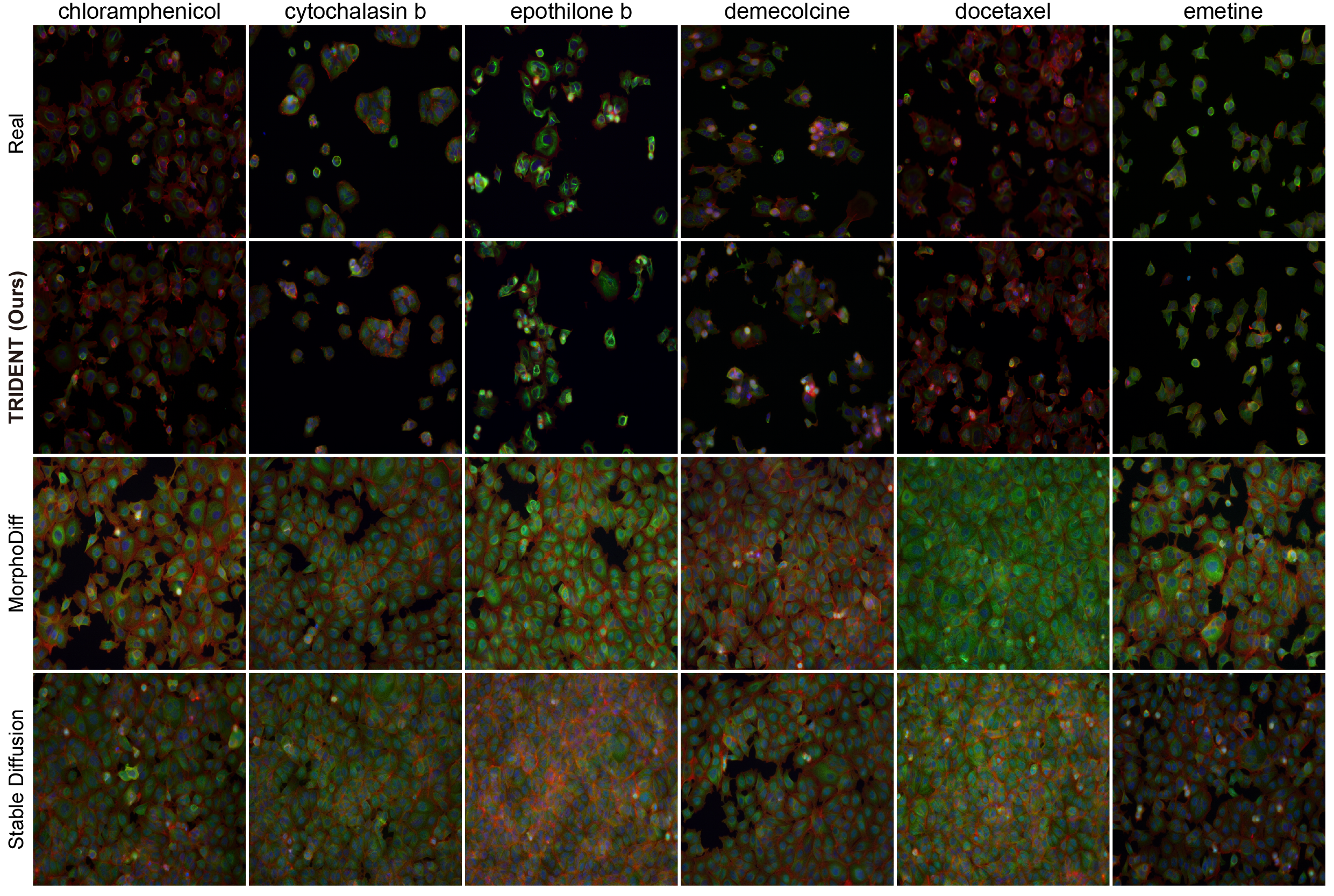}
  \vspace{-1.5em}
  \caption{Visual comparison of generated cellular morphologies under six drug perturbations. Ground-truth images (Row 1) are compared to outputs from TRIDENT (Row 2), MorphoDiff (Row 3), and Stable Diffusion (Row 4). See supplementary material for more results.}
  \label{fig3}
  \vspace{-1em}
\end{figure*}

\subsection{Comparison with State-of-the-Art Methods}
We evaluate image quality generated by TRIDENT against two state-of-the-art (SOTA) diffusion methods: MorphoDiff and a fine-tuned unconditional Stable Diffusion. All models were trained on our MorphoGene dataset for 10,000 steps.

Qualitatively, TRIDENT captures complex, drug-specific cellular patterns across six distinct perturbations (\cref{fig3}). Many of these compounds are inhibitors that impede cell proliferation (e.g., by disrupting cytoskeletal dynamics), leading to phenotypes with reduced cell density. TRIDENT's generated images are virtually indistinguishable from ground-truth, replicating unique phenotypes like changes in cell shape and population density. For instance, it uniquely generates the characteristic low cell density for cytochalasin b \cite{maclean1980mechanism,theodoropoulos1994cytochalasin}. In contrast, both baselines fail to produce specific phenotypes, collapsing to a generic, high-density monolayer and thus failing to learn the conditional guidance.

Quantitatively, we benchmark using Fréchet Inception Distance (FID) \cite{heusel2017gans}, Kernel Inception Distance (KID) \cite{binkowski2018demystifying}, and Inception Score (IS) \cite{salimans2016improved} on ID and OOD test sets (\cref{table1}). Lower scores are preferable for all three metrics. FID and KID measure distributional similarity between generated and real images (lower = higher fidelity). For IS, in this context, a lower score is preferable as it indicates that the model has learned to consistently generate the specific, constrained phenotype for a given condition, rather than an overly broad range of morphologies.

TRIDENT demonstrates a profound improvement over both baselines (\cref{table1}). On the ID test set, its FID (49.770) represents a 5- to 7-fold improvement over MorphoDiff (250.290) and Stable Diffusion (354.576). A large performance gap is also seen in the KID metric (0.013 vs. 0.248 and 0.378). Crucially, this superiority extends to the OOD task, which evaluates generalization to unseen compounds. Here, TRIDENT's FID (126.150) is more than 3-fold better than the SOTA baselines (387.135 and 393.129).

This quantitative and qualitative evidence confirms that TRIDENT's explicit modeling of the \textit{(Perturbation + RNA) $\rightarrow$ Morphology} pathway enables superior fidelity and stronger alignment to the target perturbation's phenotypic space, for both known and unseen compounds.

\begin{figure*}[t]
  \centering
  \begin{subfigure}{0.69\linewidth}
    \includegraphics[width=\linewidth]{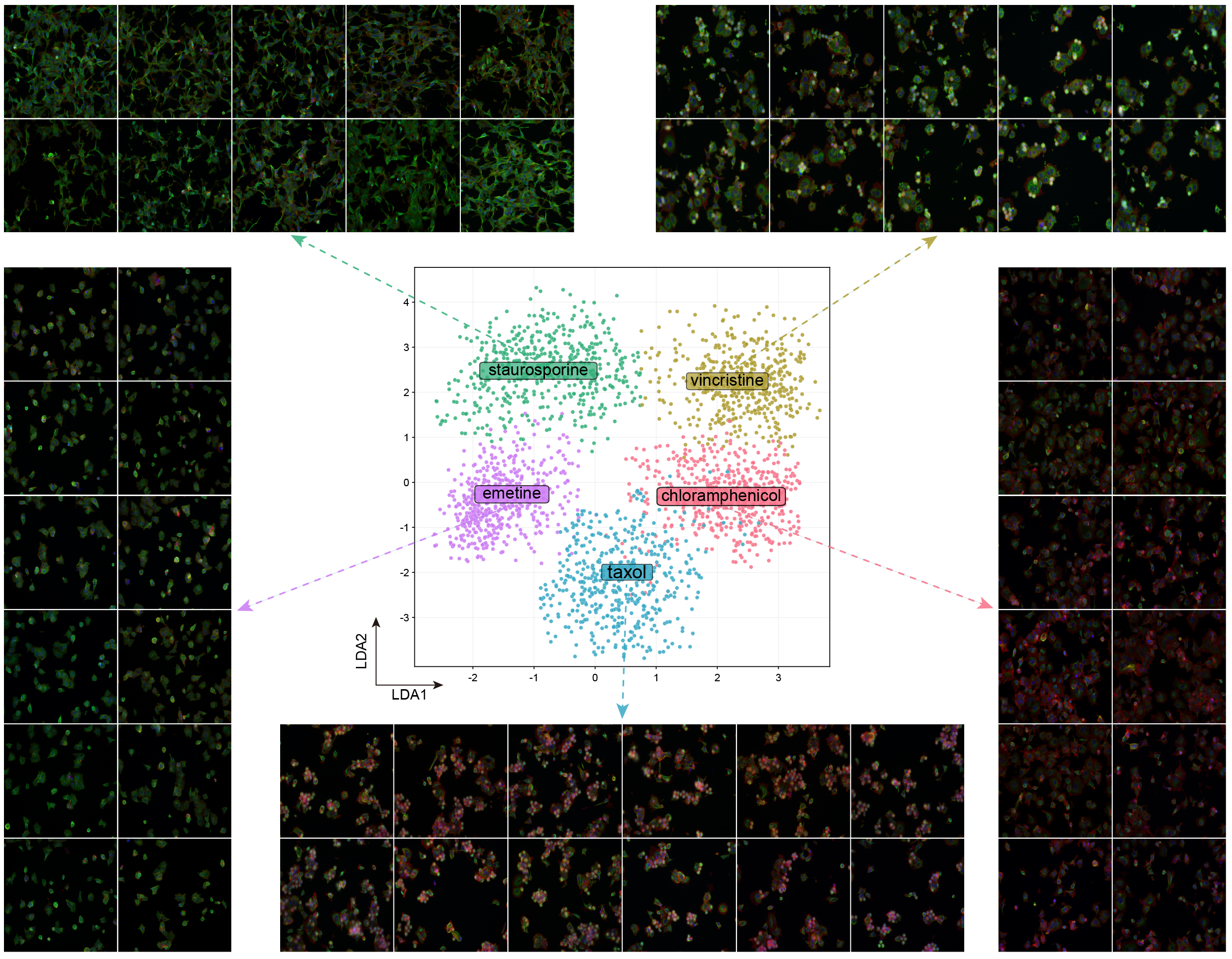}
    \caption{Phenotypic separability in LDA embedding space.}
    \label{fig4a}
  \end{subfigure}
  \begin{subfigure}{0.30\linewidth}
    \includegraphics[width=\linewidth]{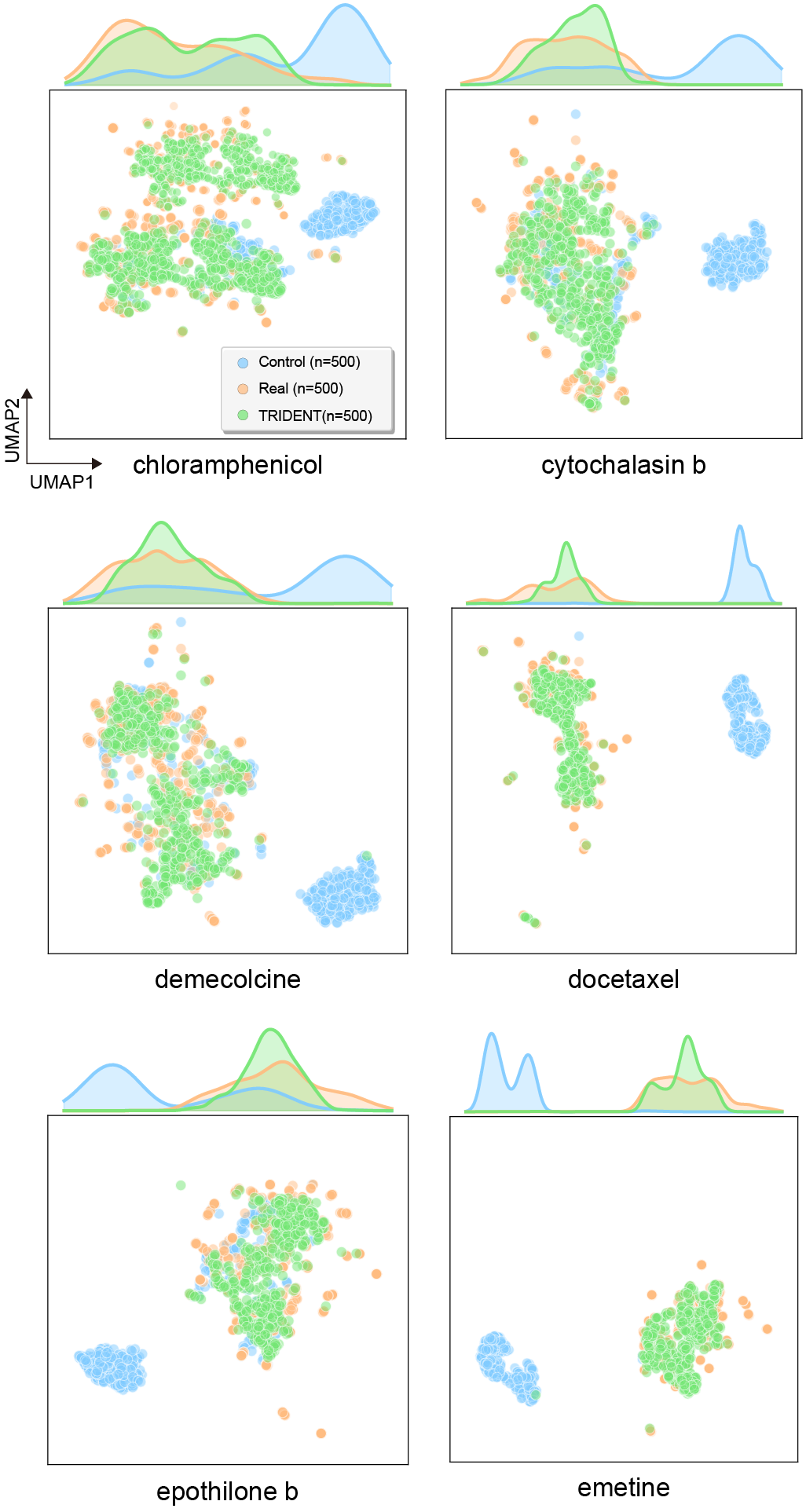}
    \caption{UMAP embedding space.}
    \label{fig4b}
  \end{subfigure}
  \par\medskip
  \begin{subfigure}[t]{\linewidth}
    \includegraphics[width=\linewidth]{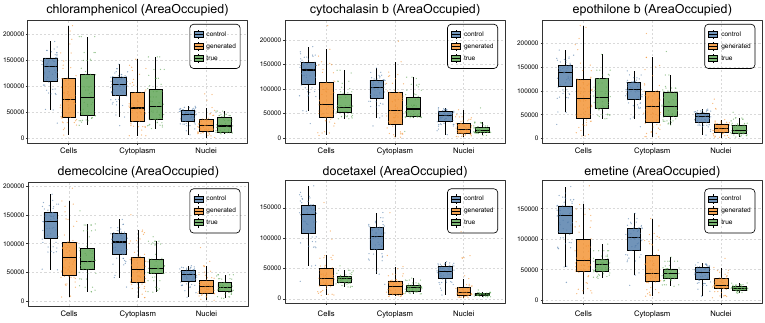}
    \caption{Quantitative comparison of morphological features.}
    \label{fig4c}
  \end{subfigure}
  \caption{TRIDENT captures biologically interpretable signatures in embedding and feature space. (a) ViT embeddings of generated images form distinct, MOA-specific clusters in LDA space, with representative images shown for each cluster. (b) UMAP visualization confirms high distributional alignment between generated (green) and real (orange) images, which are both separate from the control (blue) population. (c) Quantitative CellProfiler analysis shows that \textit{AreaOccupied} feature distributions of generated and real images are highly similar and distinct from control across all cellular compartments.}
  \label{fig4-all}
  \vspace{-1em}
\end{figure*}

\subsection{High Dimensional Feature Analysis}
Building upon the validation of high-fidelity image generation, we next assess if TRIDENT captures biologically interpretable morphological signatures in high-dimensional embedding and feature spaces. We project generated images of compounds with diverse Mechanism of Action (MOA) into a pre-trained Vision Transformer (ViT) \cite{dosovitskiy2020image} embedding space and visualize them via Linear Discriminant Analysis (LDA) \cite{balakrishnama1998linear} (\cref{fig4a}). The results reveal that generated images form distinct, tightly-clustered groups corresponding to their specific drug perturbations. For instance, the filamentous morphology generated for the kinase inhibitor staurosporine \cite{karaman2008quantitative,tanramluk2009origins,chae2000molecular,bruno1992different} is clearly separated from the sparse, rounded-cell phenotype predicted for the protein synthesis inhibitor emetine \cite{gupta1977molecular,rhoads1985emetine,gupta1980structural}, confirming the model's ability to capture distinct, MOA-specific biological programs.

Having established that TRIDENT's representations are separable by MOA, we next evaluate their fidelity to the ground-truth. A UMAP visualization \cite{mcinnes2018umap} (\cref{fig4b}) shows a striking alignment: embeddings from TRIDENT-generated and real images are tightly intermingled and occupy a shared manifold, separate from the control population. This demonstrates that generated morphologies are biologically correct and virtually indistinguishable from real data in this embedding space. This finding is further substantiated by a quantitative analysis of interpretable cytological measurements using CellProfiler \cite{stirling2021cellprofiler} (\cref{fig4c}). The analysis of key features like \textit{AreaOccupied} shows a remarkable alignment between the feature distributions of the generated and true images across the whole cell, cytoplasm, and nucleus. Both consistently exhibit a significant shift away from the control population, proving that TRIDENT accurately recapitulates the nuanced, multi-scale morphological changes induced by drug perturbations.

\begin{figure*}[t]
  \centering
  \begin{subfigure}{\linewidth}
    \includegraphics[width=\linewidth]{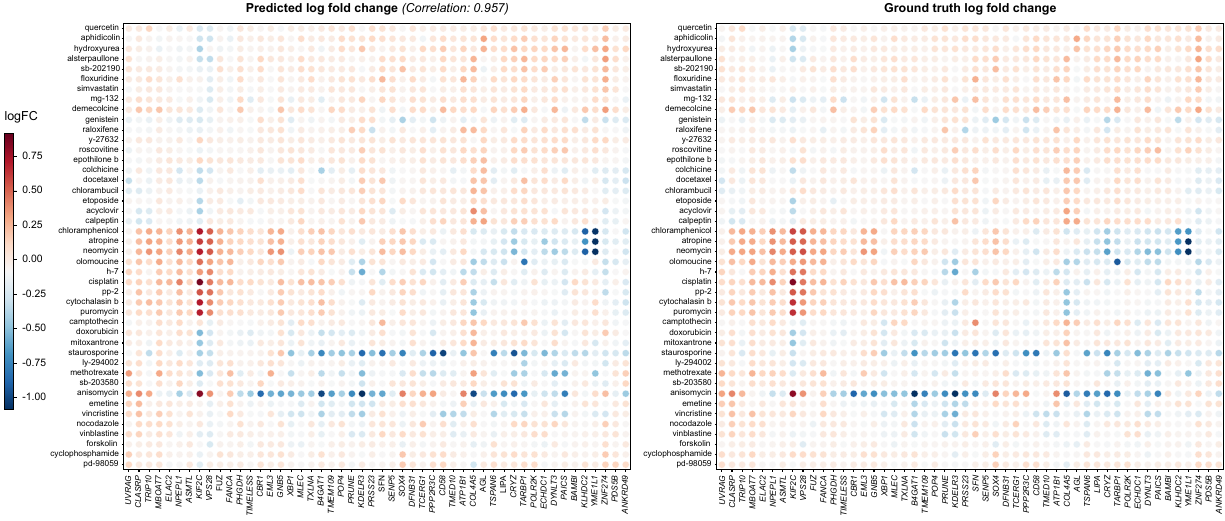}
    \caption{Prediction of transcriptomic log fold change.}
    \label{fig5a}
  \end{subfigure}
  \par\medskip
  \begin{subfigure}[t]{0.51\linewidth}
    \includegraphics[width=\linewidth]{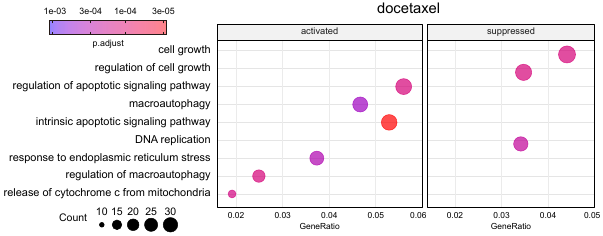}
    \caption{Functional enrichment analysis for docetaxel.}
    \label{fig5b}
  \end{subfigure}
  \hfill
  \begin{subfigure}[t]{0.48\linewidth}
    \includegraphics[width=\linewidth]{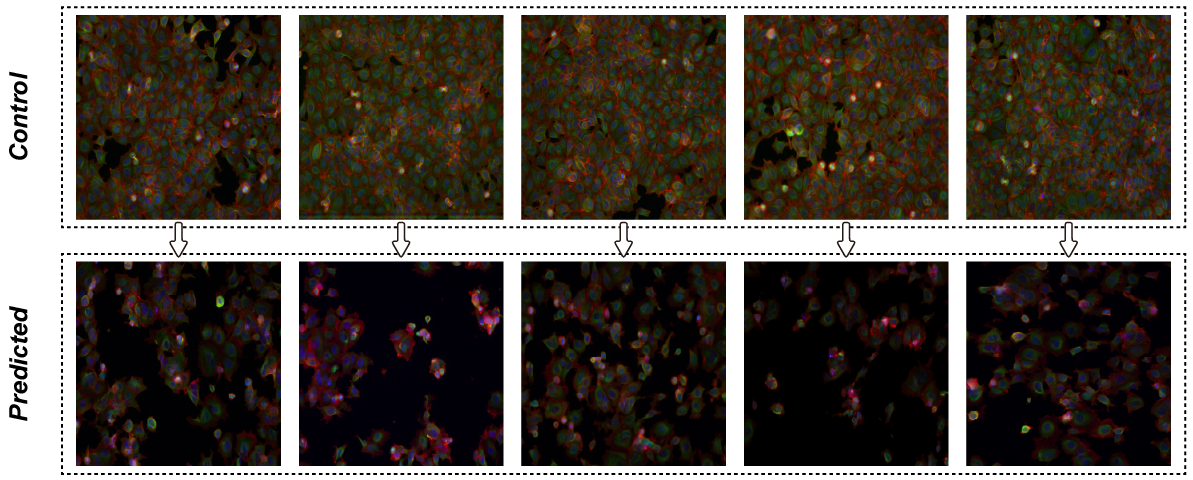}
    \caption{Predicted morphological response to docetaxel.}
    \label{fig5c}
  \end{subfigure}
  \caption{TRIDENT learns the association between transcriptome and morphology. (a) Heatmaps comparing predicted (left) versus ground-truth (right) gene expression log fold changes for 44 compounds. (b) Functional enrichment analysis of model-predicted genes for docetaxel identifies pathways consistent with its known MOA. (c) Visual comparison of TRIDENT-predicted morphology for docetaxel (bottom) versus control (top), correctly capturing the phenotype of reduced cell density.}
  \label{fig5-all}
  \vspace{-1em}
\end{figure*}

\subsection{Cross-Modal Validation}
We validate if TRIDENT learns the true biological mapping between the molecular state (RNA) and the physical phenotype (morphology). We first assess the accuracy of the post-perturbation transcriptome ($G_{post}$) predicted by the Transcription-Drug Condition Module, and then its consistency with the final generated morphology.

We evaluated the global accuracy of the predicted gene expression profiles. For 44 ID compounds, the Z-scored predicted log fold change (LFC) patterns demonstrate a striking similarity to the ground-truth patterns (\cref{fig5a}). This is confirmed quantitatively by a high Pearson correlation of 0.957, indicating the module generates highly accurate and biologically faithful transcriptional profiles. To further validate the biological relevance of these predictions, we perform a case study using docetaxel \cite{lyseng2005docetaxel,clarke1999clinical,tannock2004docetaxel}, a well-characterized compound. We partition the genes into upregulated (activated) and downregulated (suppressed) sets based on their LFC generated by TRIDENT. Functional enrichment analysis on these genes aligns perfectly with docetaxel's known MOA as a mitotic inhibitor that promotes cell cycle arrest and apoptosis (\cref{fig5b}). Suppressed genes are significantly enriched for terms like \textit{regulation of cell growth}' and \textit{DNA replication}, while activated genes are enriched for \textit{regulation of apoptotic signaling pathway}. This demonstrates the predicted transcriptome is functionally and mechanistically correct.

Finally, we investigate the critical cross-modal link: whether this predicted biological program (i.e., suppressed growth, induced apoptosis) is reflected in the generated morphology. The biological functions identified in \cref{fig5b} strongly imply a phenotype characterized by reduced cell proliferation and increased cell death, which manifests as a significant decrease in cell density. \cref{fig5c} provides a direct visual comparison, showing that TRIDENT-generated images for docetaxel clearly exhibit a sparse cell population. This precisely matches the expected morphological outcome and aligns with our quantitative CellProfiler analysis (\cref{fig4c}).

This evidence demonstrates that TRIDENT successfully learns the complex, cross-modal association between transcriptome and morphology, where functionally-correct RNA predictions actively and accurately guide the synthesis of the correct corresponding cellular phenotype.

\begingroup
\setlength{\tabcolsep}{4pt}
\renewcommand{\arraystretch}{1.35}
\begin{table}[t]
  \centering
  \caption{Ablation study results. Performance comparison of the full TRIDENT model against a variant without the RNA conditioning module on ID and OOD test sets.}
  \label{table2}
  \begin{adjustbox}{width=\columnwidth}
  \begin{tabular}{l|ccc|ccc}
  \toprule\toprule
  \multirow{2}{*}{\centering\bfseries\large Methods} &
    \multicolumn{3}{c|}{\textbf{In-Distribution}} &
    \multicolumn{3}{c}{\textbf{Out-of-Distribution}} \\
  \cmidrule(lr){2-4}\cmidrule(lr){5-7}  
  & FID$\downarrow$ & KID$\downarrow$ & IS$\downarrow$ & FID$\downarrow$ & KID$\downarrow$ & IS$\downarrow$ \\
  \midrule\midrule
    TRIDENT & 49.770 & 0.013 & 2.240  & 126.150 & 0.222 & 2.523 \\
    TRIDENT (w/o RNA) & 115.770 & 0.132 & 2.381 & 194.239 & 0.293 & 2.639 \\
    \bottomrule\bottomrule
  \end{tabular}
  \end{adjustbox}
  \vspace{-1em}
\end{table}
\endgroup

\subsection{Ablation Studies}
To demonstrate the importance of RNA expression as a critical intermediate for accurate morphology synthesis, we conduct an ablation study. We create a variant, TRIDENT (w/o RNA), which bypasses the \textbf{Transcription-Drug Condition Module} and uses only the drug embedding $D$ to condition the diffusion model. This models a direct \textit{Perturbation $\rightarrow$ Morphology} pathway. As shown in \cref{table2}, the full TRIDENT framework significantly outperforms this ablated model. On the ID test set, TRIDENT's FID (49.770) is more than 2.3-fold better than the ablated model's (115.770). The performance gap is even more pronounced in the KID metric (0.013 vs. 0.132), an order of magnitude improvement. This demonstrates that the drug information $D$ alone is an insufficient condition. The latent vector $z$, which integrates both perturbation and transcriptional state, provides a far richer signal. The full model's superior performance on the OOD set (FID 126.150 vs. 194.239) further confirms that explicitly modeling the \textit{(Perturbation + RNA) $\rightarrow$ Morphology} pathway is essential for high fidelity and accuracy.

%% file: sec/5_conclusion_limitation.tex
\section{Conclusion}
In this work, we address a critical gap in cellular modeling: the cross-modal mapping from transcriptome to phenotype. We introduce TRIDENT, a cascaded generative framework modeling the complete tripartite relationship between perturbation, gene expression, and cellular morphology. TRIDENT generates high-fidelity, MOA-specific morphologies, significantly outperforming SOTA methods quantitatively and generalizing better to unseen compounds. We validated that the high fidelity of the generated cellular morphology is guided by functionally-correct transcriptomes. Ablation studies confirmed this intermediate RNA state is essential for high-fidelity generation. By mechanistically linking transcriptional state to morphological outcome, TRIDENT provides a foundational component essential for realizing the AIVC vision, enabling simulations where molecular events causally orchestrate cellular form.

\vspace{0.5em}
\noindent\textbf{Limitations.} A primary limitation is our MorphoGene dataset, which uses a single cell line (MCF7) and bulk L1000 profiles. This limits cross-cell-type generalization and overlooks cell-to-cell heterogeneity. Future work should prioritize richer, multi-modal datasets, ideally pairing single-cell transcriptomics with imaging across diverse cell lines. Such data will enable more generalizable models, including zero-shot, cross-cell-line prediction and the pre-training of large-scale foundational models, marking the next step toward a predictive virtual cell.

%% file: sec/X_suppl.tex
\clearpage
\appendix
\setcounter{figure}{0} 
\setcounter{section}{0} 
\setcounter{figure}{0} 
\renewcommand{\thefigure}{A\arabic{figure}} 

\twocolumn[ 
  \begin{@twocolumnfalse}
    \begin{center}
      {\Large \bfseries Supplemental Material: \\
TRIDENT: A Trimodal Cascade Generative Framework for Drug and RNA-Conditioned Cellular Morphology Synthesis \par}
      \vspace{2em}
      {\large Rui Peng$^{1,2}\textsuperscript{\#}$~
    Ziru Liu$^5\textsuperscript{\#}$~
    Lingyuan Ye$^{6,7}$~
    Yuxing Lu$^{1}$~
    Boxin Shi$^{3,4}\textsuperscript{*}$~
    Jinzhuo Wang$^{1}$\textsuperscript{*}} \\
    \vspace{0.5em}
    {\small \textsuperscript{1} Department of Big Data and Biomedical AI, College of Future Technology, Peking University\\
    \small \textsuperscript{2} Center for BioMed-X Research, Academy for Advanced Interdisciplinary Studies, Peking University\\
    \small \textsuperscript{3} State Key Laboratory of Multimedia Information Processing, School of Computer Science, Peking University\\ 
    \small \textsuperscript{4} National Engineering Research Center of Visual Technology, School of Computer Science, Peking University\\
    \small \textsuperscript{5} Yuanpei College, Peking University\qquad \textsuperscript{6} School of Life Sciences, Tsinghua University\\
    \small \textsuperscript{7} Peking University-Tsinghua University-National Institute of Biological Sciences Joint Graduate Program (PTN), Tsinghua University\\
    \small{\texttt{\{pengrui, lzr, luyx\}@stu.pku.edu.cn~~~yely23@mails.tsinghua.edu.cn}}\\
    \small{\texttt{shiboxin@pku.edu.cn~~~wangjinzhuo@pku.edu.cn}}}
    \vspace{2em}
    \end{center}
  \end{@twocolumnfalse}
]

\begingroup
  \renewcommand\thefootnote{}
  \footnotetext{\# Equal contribution.\quad * Corresponding authors.}
  \addtocounter{footnote}{-1}
\endgroup

\section{Implementation Details} 
All models are implemented in PyTorch and trained on a high-performance computing cluster equipped with eight NVIDIA A100 GPUs, each with 80GB of memory. The core of our Morphology Generation Module is a Diffusion Transformer architecture. This transformer is configured with 28 layers, a hidden dimension of 1152, and 16 attention heads. The complete TRIDENT framework is trained end-to-end for a total of 100,000 steps. We employ the AdamW optimizer with a constant learning rate of 1e-4. A global batch size of 32 is used, distributed across the eight GPUs. All Cell Painting images are processed at a resolution of $512\times512$ pixels. The total training process for the final model takes approximately four days.

\section{Cellprofiler Feature Construction} 
To derive quantitative descriptors of cellular phenotype, we construct a bespoke analysis workflow using CellProfiler (v5.0). This pipeline is engineered to process each Cell Painting image and output a single, comprehensive feature vector summarizing its morphological characteristics.

The workflow's core is a three-step segmentation process to delineate cellular structures. First, nucleus are identified as primary objects from the blue (DNA) channel. Next, cell boundaries are segmented as secondary objects by propagating outwards from the identified nucleus, using the green channel to define the cell periphery. Finally, the cytoplasm is defined as a tertiary region, calculated by subtracting the nuclear mask from the corresponding cell mask. A quality control step is integrated to discard all objects touching the image border, ensuring that all downstream measurements are derived from complete, intact cells.

Following segmentation, a comprehensive suite of CellProfiler modules is executed to extract measurements from all three compartments (nucleus, cytoplasm, and cells) across all channels. The extracted features include morphological descriptors of size and shape, such as area, perimeter, major and minor axis lengths, eccentricity, and solidity. Furthermore, statistics on pixel intensity distribution like mean, median, standard deviation, median absolute deviation, and quartiles are computed. Finally, the pipeline captures relational metrics, such as spatial relationships between cells, inter-channel signal correlations, and radial intensity distributions.

To generate a single profile for each image, these per-object measurements are aggregated by calculating their mean, median, and standard deviation across all valid cells. This process yields a final, high-dimensional profile of 6,345 distinct morphological features for each image, providing a detailed quantitative fingerprint of the cellular phenotype in response to perturbation.

\newpage
\section{TRIDENT Algorithm}
\begin{algorithm}[h]
\caption{TRIDENT Framework: Training Procedure}
\label{alg:trident_training}
\begin{algorithmic}[1]
\Require 
    Training data $\mathcal{D} = \{(G_{pre}^{(i)}, D^{(i)}, I^{(i)}, G_{post}^{(i)})\}$
\Require 
    Pre-trained image VAE ($\mathcal{E}_{image}, \mathcal{D}_{image}$)
\Require 
    Diffusion timesteps $T$, variance schedule $\{\beta_t\}_{t=1}^T$ (and $\alpha_t, \bar{\alpha}_t$)
\Ensure 
    Trained parameters $\Theta = \{\phi, \psi, \theta, \gamma \}$

\Statex
\State Initialize VAE parameters $\phi$ (for $\mathcal{E}_{rna}, \mathcal{E}_{drug}$), $\psi$ (for $\mathcal{D}_{perturb}$)
\State Initialize Denoising Transformer $f_{\theta}$ parameters $\theta$
\State Initialize Time Embedding parameters $\gamma$ for $\mathcal{E}_{time}$

\For{each epoch $e = 1, \dots, E_{max}$}
    \For{each batch $(G_{pre}, D, I, G_{post}) \sim \mathcal{D}$}
        \State $\mat{X}_{rna} \leftarrow \mathcal{E}_{rna}(G_{pre})$, $\mat{X}_{drug} \leftarrow \mathcal{E}_{drug}(D)$
        \State $[\vect{\mu}_z, \log \vect{\sigma}^2_z] \leftarrow \mathcal{E}_{perturb}([\mat{X}_{rna}, \mat{X}_{drug}])$
        \State $\vect{\epsilon}_z \sim \mathcal{N}(\vect{0}, \mat{I})$
        \State $\vect{z} \leftarrow \vect{\mu}_z + \vect{\sigma}_z \odot \vect{\epsilon}_z$ 
        \State $[\vect{\mu}_{G_{post}}, \log \vect{\sigma}^2_{G_{post}}] \leftarrow \mathcal{D}_{perturb}(\vect{z})$
        \State $\mathcal{L}_{recon} \leftarrow \E_{q_{\phi}}[-\log p_{\psi}(G_{post} | \vect{z})]$ 
        \State $\mathcal{L}_{KL} \leftarrow \KL(q_{\phi}(\vect{z} | G_{pre}, D) \,||\, p(\vect{z}))$ 
        \State $\mathcal{L}_{VAE} \leftarrow \mathcal{L}_{recon} + \mathcal{L}_{KL}$
        
        \State $\mat{X}^0_{image} \leftarrow \mathcal{E}_{image}(I)$ 
        \State $t \sim \mathcal{U}(\{1, \dots, T\})$ 
        \State $\vect{\epsilon} \sim \mathcal{N}(\vect{0}, \mat{I})$ 
        \State $\mat{X}^t_{image} \leftarrow \sqrt{\bar{\alpha}_t}\mat{X}^0_{image} + \sqrt{1-\bar{\alpha}_t}\vect{\epsilon}$ 
        \State $\mat{X}_{time} \leftarrow \mathcal{E}_{time}(t)$ 
        \State $\mat{X}_{condition} \leftarrow \vect{z} + \mat{X}_{time}$ 
        \State $\vect{\epsilon}_{\theta} \leftarrow f_{\theta}(\mat{X}^t_{image}, \mat{X}_{condition})$ 
        \State $\mathcal{L}_{LDM} \leftarrow || \vect{\epsilon} - \vect{\epsilon}_{\theta} ||^2$ 
        
        \State $\mathcal{L}_{TRIDENT} \leftarrow \mathcal{L}_{VAE} + \mathcal{L}_{LDM}$ 
        \State Update parameters $\phi, \psi, \theta, \gamma $ using $\nabla \mathcal{L}_{TRIDENT}$ 
        
    \EndFor
\EndFor
\State \textbf{return} Trained parameters $\Theta = \{\phi, \psi, \theta, \gamma\}$
\end{algorithmic}
\end{algorithm}


\begin{algorithm}[h]
\caption{TRIDENT Framework: Inference Procedure}
\label{alg:trident_inference}
\begin{algorithmic}[1]
\Require 
    Input $G_{pre}$, $D$
\Require 
    Trained parameters $\Theta = \{\phi, \psi, \theta, \gamma \}$
\Require
    Pre-trained image VAE ($\mathcal{E}_{image}, \mathcal{D}_{image}$)
\Require
    Diffusion timesteps $T$ and schedule $\{\beta_t\}_{t=1}^T$
\Ensure 
    Generated Image $\hat{I}$

\State $\mat{X}_{rna} \leftarrow \mathcal{E}_{rna}(G_{pre})$
\State $\mat{X}_{drug} \leftarrow \mathcal{E}_{drug}(D)$
\State $[\vect{\mu}_z, \log \vect{\sigma}^2_z] \leftarrow \mathcal{E}_{perturb}([\mat{X}_{rna}, \mat{X}_{drug}])$
\State $\vect{\epsilon}_z \sim \mathcal{N}(\vect{0}, \mat{I})$
\State $\vect{z} \leftarrow \vect{\mu}_z + \vect{\sigma}_z \odot \vect{\epsilon}_z$ 

\State $\hat{\mat{X}}^T_{image} \sim \mathcal{N}(\vect{0}, \mat{I})$ 
\For{$t = T$ \textbf{down to} $1$}
    \State $\mat{X}_{time} \leftarrow \mathcal{E}_{time}(t)$
    \State $\mat{X}_{condition} \leftarrow \vect{z} + \mat{X}_{time}$
    \State $\vect{\epsilon}_{\theta} \leftarrow f_{\theta}(\hat{\mat{X}}^t_{image}, \mat{X}_{condition})$ 
    \State $\vect{\epsilon}' \sim \mathcal{N}(\vect{0}, \mat{I})$ if $t > 1$, else $\vect{\epsilon}' \leftarrow \vect{0}$
    \State $\hat{\mat{X}}^{t-1}_{image} \leftarrow \frac{1}{\sqrt{\alpha_t}}\left( \hat{\mat{X}}^t_{image} - \frac{\beta_t}{\sqrt{1-\bar{\alpha}_t}} \vect{\epsilon}_{\theta} \right) + \sigma_t \vect{\epsilon}'$ 
\EndFor

\State $\hat{I} \leftarrow \mathcal{D}_{image}(\hat{\mat{X}}^0_{image})$ 
\State \textbf{return} $\hat{I}$
\end{algorithmic}
\end{algorithm}

\section{Additional Comparison Results}
\begin{figure*}[t]
  \centering
  \includegraphics[width=\linewidth]{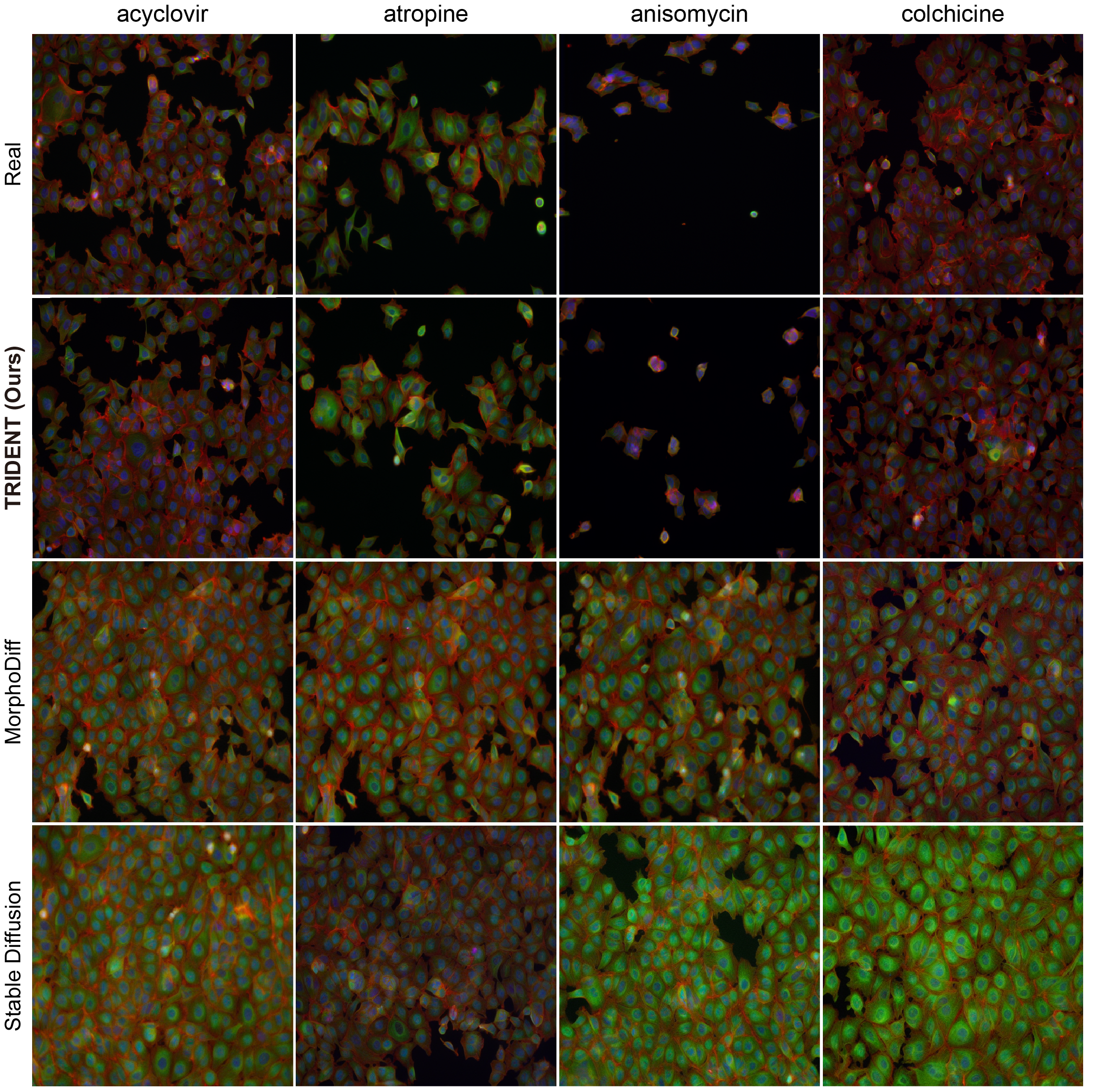}
  \caption{Additional visual comparison of generated cellular morphologies. Ground-truth images (Row 1) are compared to outputs from TRIDENT (Row 2), MorphoDiff (Row 3), and Stable Diffusion (Row 4).}
  \label{fig3}
\end{figure*}

\begin{figure*}[h]
  \centering
  \includegraphics[width=\linewidth]{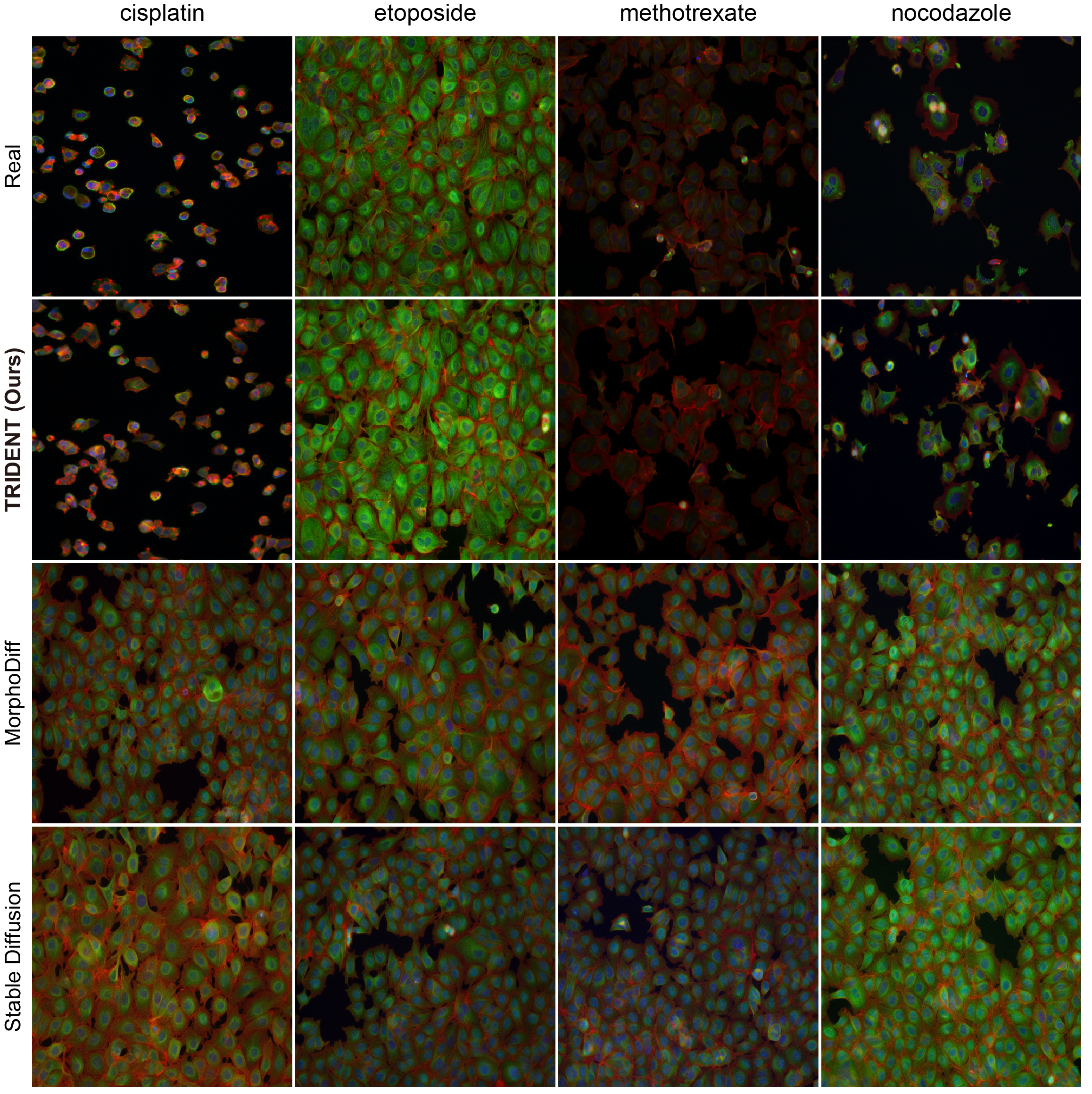}
  \caption{Additional visual comparison of generated cellular morphologies. Ground-truth images (Row 1) are compared to outputs from TRIDENT (Row 2), MorphoDiff (Row 3), and Stable Diffusion (Row 4).}
  \label{fig3}
\end{figure*}

\begin{figure*}[h]
  \centering
  \includegraphics[width=\linewidth]{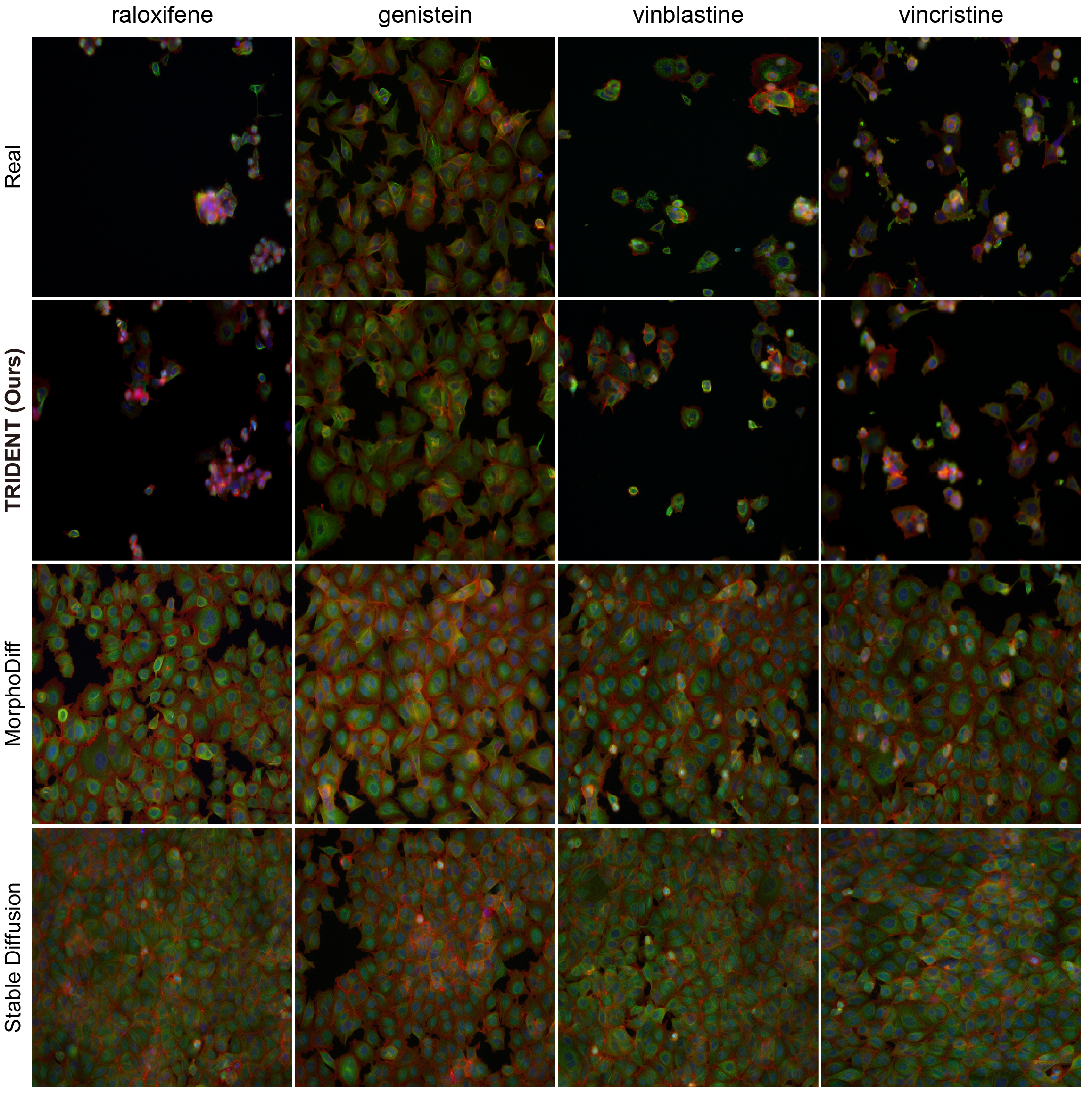}
  \caption{Additional visual comparison of generated cellular morphologies. Ground-truth images (Row 1) are compared to outputs from TRIDENT (Row 2), MorphoDiff (Row 3), and Stable Diffusion (Row 4).}
  \label{fig3}
\end{figure*}